\documentclass[10pt,twocolumn,letterpaper]{article}

\usepackage{wacv}
\usepackage{times}
\usepackage{epsfig}
\usepackage{graphicx}
 \usepackage{subcaption}
\usepackage{amsmath}
\usepackage{amssymb}
\usepackage{multirow}
 \usepackage{adjustbox}
 \usepackage{hhline}
 \usepackage{cite}

\wacvfinalcopy 

\def\wacvPaperID{****} 

\ifwacvfinal\fi
\begin{document}

\title{\textit{Two Stream LSTM} : A Deep Fusion Framework for Human Action Recognition}

\author{Harshala Gammulle \hspace{1cm} Simon Denman \hspace{1cm} Sridha Sridharan \hspace{1cm} Clinton Fookes\\
\\Image and Video Laboratory, Queensland University of Technology (QUT), Brisbane, QLD, Australia\\
{\tt\small pranaliharshala.gammulle@hdr.qut.edu.au, \{s.denman, s.sridharan, c.fookes\}@qut.edu.au}
}

\maketitle
\ifwacvfinal\thispagestyle{empty}\fi

\begin{abstract}
   In this paper we address the problem of human action recognition from video sequences. Inspired by the exemplary results obtained via automatic feature learning and deep learning approaches in computer vision, we focus our attention towards learning salient spatial features via a convolutional neural network (CNN) and then map their temporal relationship with the aid of Long-Short-Term-Memory (LSTM) networks. Our contribution in this paper is a deep fusion framework that more effectively exploits spatial features from CNNs with temporal features from LSTM models. We also extensively evaluate their strengths and weaknesses. We find that by combining both the sets of features, the fully connected features effectively act as an attention mechanism to direct the LSTM to interesting parts of the convolutional feature sequence. The significance of our fusion method is its simplicity and effectiveness compared to other state-of-the-art methods. The evaluation results demonstrate that this hierarchical multi stream fusion method has higher performance compared to single stream mapping methods allowing it to achieve high accuracy outperforming current state-of-the-art methods in three widely used databases: UCF11, UCFSports, jHMDB.
\end{abstract}


\section{Introduction}

	Recognition of human actions has gained increasing interest within the research community due to its applicability to areas such as surveillance, human machine interfaces, robotics, video retrieval and sports video analysis. It is still an open research challenge, especially in real world scenarios with the presence of background clutter and occlusions.

\begin{figure}[h]
        \centering
        	\includegraphics[width=0.3\textwidth]{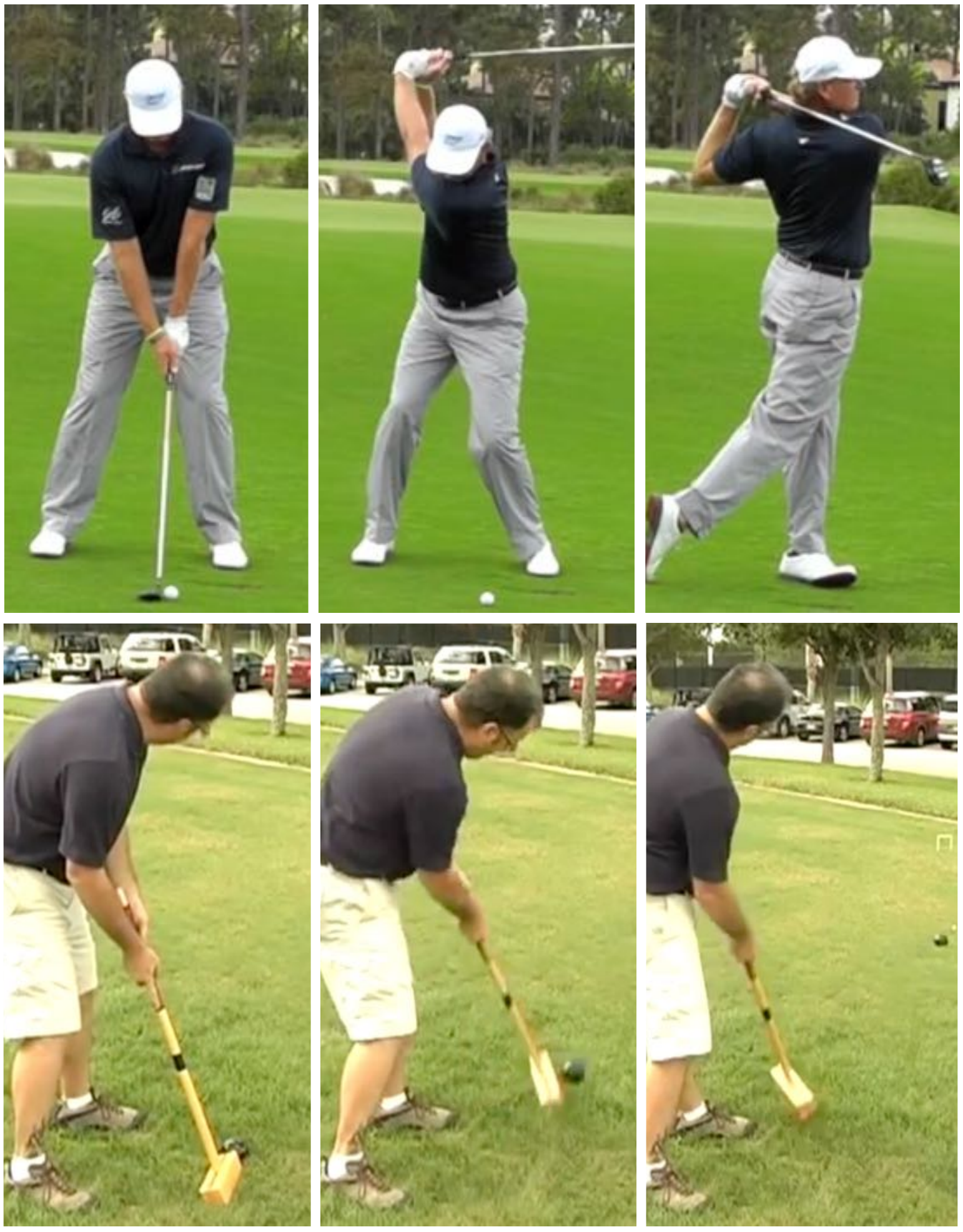}
	\caption{Golf Swing vs Croquet Swing. While the two actions are visibly similar (green background, similar athlete pose), the different motion patterns allow the actions to be separated.}
	\label{fig:fig3}
       
        \end{figure}

	It is widely regarded \cite{Simonyan2014, Jeff2015}, that the strongest opportunity to effectively recognise human actions from video sequences is to exploit the presence of both spatial and temporal information. Although, it is still possible to obtain meaningful outputs using only the spatial domain  due to other cues that are present in the scene, in some instances using only a single domain can lead to confusion and errors. For example, an action such as `golf swing' has some unique properties when compared to other action classes. Backgrounds are typically green and the objects `golf club' and `golf ball' are present. Whilst these attributes can help distinguish between actions such as `golf swing' and `ride bike' where a very different arrangement of objects is present, they are less helpful when comparing actions such as `golf swing' and `croquet swing', which are visually similar. In such situations, the temporal domain of the sequence plays an important role as it provides information on the temporal evolution of the spatial features. As shown in Figure \ref{fig:fig3} the motion patterns observed in `golf swing' are different from `croquet swing', despite other visual similarities.

\subsection{The Proposed Approach}
	
	When considering real world scenarios, it is preferable to have an automatic feature learning approach, as handcrafted features will only capture abstract level concepts that are not truly representative of the action classes of interest. Hierarchical (or deep) learning of features in an unsupervised  way will capture more complex concepts, and has been shown to be effective in other tasks such as audio classification \cite{deepaudio}, image classification \cite{deepImage1,deepImage2,deepImage3} and scene classification \cite{deepScene}. Existing approaches have looked at the combination of deep features with sequential models like LSTMs, however, it is still not well understood why and how this combination should occur. We show in this paper how the particular combination of both convolutional and fully connected activations from a CNN combined with a two stream LSTM can not only exceed the state-of-the-art performance, but can do so through a simple manner with improved computational efficiency. Furthermore, we uncover the strengths and weaknesses of the fusion approach to highlight how the fully connected features can direct the attention of the LSTM network to the most relevant parts of the convolutional feature sequences. Our model can be considered a simpler model compared to related previous works such as \cite{Simonyan2014, actiontubes, Jeff2015, Sharma2015, Yao2015}, yet our proposed approach out performs current state of the art methods in several challenging action databases.       
	
 	\vspace{-2 mm}
  In Section 2 we describe related efforts and explain how our proposed method differs from these. Section 3 describes our  action recognition model.  In Section 4 we describe the databases used, discuss the experiments performed and results; and Section 5 concludes the paper. 


\section{Relation to previous efforts}    

In the field of human action recognition, hitherto proposed approaches can be broadly categorised as static or video based. In static approaches background information or the position of human body/body parts \cite{Delaitre2010,Raja2011} is used to aid the recognition. However, video based approaches are seen to be more appropriate as they represent both the spatial and temporal aspects of the action \cite{Simonyan2014}.

	When considering video based approaches, most studies have focused on extracting effective handcrafted features that aid in classifying actions well. For instance in the work by Abdulmunem et al. \cite{Abdulmunem2016}, they have used hand crafted features by extracting saliency guided 3D-SIFT-HOOF (SGSH) features which are concatenated 3D SIFT and histogram of oriented optical flow (HOOF) features; these are subsequently fed to a Support Vector Machine (SVM) after encoding them using the bag of visual words approach. Due to the drawback of such handcrafted features capturing information only at an abstract level, recent approaches have tended to focus on utilising unsupervised feature learning methods. For instance, the method proposed by Simonyan et al. \cite{Simonyan2014} uses two separate Convolutional Neural Networks (CNN) to learn features from both the spatial and temporal domains of the video. The final output is obtained through two different fusion methods; averaging and training a multi-class linear SVM using the soft-max scores as features. This idea of a dual network also been used by Gkioxari et al. \cite{actiontubes}  in their action tubes approach, using optical flow to locate salient image regions that contain motion which are fed as input to the CNN rather than the entire video frame. But the use of multiple replicated networks adds complexity and it requires the training of a network for each domain. In an earlier work by Ji et al. \cite{Ji2013}, a single CNN model is used to learn from features including grey values, gradients and optical flow magnitudes which have been obtained from the input image using a `hard wired' layer. But the incorporation of hand-crafted features such as in this case can make the automatic learning approach incomplete.

		 In addition to unsupervised feature learning using CNNs, modelling sequences by combining CNNs with Recurrent Neural Networks (RNN) has become increasingly popular among the computer vision community. In \cite{Jeff2015} Donahue et al. introduced a long term recurrent convolutional network which extracts the features from a 2D CNN and passes those through an LSTM network to learn the sequential relationship between those features. In contrast to our proposed approach, the authors in \cite{Jeff2015} formulated the problem as a sequence-to-sequence prediction, where they predict an action class for every frame of the video sequence. The final prediction is taken by averaging the individual predictions. This idea is further extended by Baccouche et al. in \cite{Baccouche2011} where they utilise a 3D CNN rather than a 2D CNN network to extract features. They feed 9 frame sub video sequences from the original video sequence and the extracted spatio-temporal features are passed through a LSTM to predict the relevant action class for that 9 frame sub sequence similar to \cite{Jeff2015}. The final prediction is decided by averaging the individual predictions. In both of these approaches a dense convolutional layer output is extracted and passed through the LSTM network to generate the relevant classification.  

		 In order to combine LSTMs with DCNNs, deep features need to be extracted from the DCNN as input for the LSTM. In a study by Zhu et al. \cite{Zhu2016}, they have used a method called hyper-column features for a facial analysis task. This concept of hyper-columns is based on extracting several earlier CNN layers and several later layers, instead of extracting only the last dense layer of the CNN. Such an approach is proposed as the later fully connected layers only include semantic information and no spatial information; while the earlier convolution layers are richer in spatial information, though without the semantic details. However, these type of methods are more relevant for tasks where location information is important. Typically for action recognition, we are more concerned with the location and rotation invariance of extracted features, and such properties are not as well supported by the features provided by earlier CNN layers.

	More recently in \cite{Sharma2015,Yao2015}, authors have shown that LSTM based attention models are capable of capturing where the model should pay attention to in each frame of the video sequence when classifying the action. In \cite{Karpathy2014} Karpathy et al. used a multi resolution approach and fixed the attention to the centre of the frame. In \cite{Sharma2015}, the authors utilised a soft attention mechanism, which is trained using back propagation, in order to have dynamically varying attention throughout the video sequence. Learning such attention weights through back propagation is an exhaustive task as we have to check all possible input/output combinations. Furthermore, under evaluation results in \cite{Sharma2015} authors have shown that if attention is given to an incorrect region of the frame, it may lead to misclassification.

	In contrast to existing approaches that combine DCNNs and LSTMs by taking a single dense layer from the DCNN as input to the LSTM, we consider a multi-stream approach where information from both final convolutional layer and first fully connected layer are considered. In the following sections we explain the rationale for this choice and experimentally demonstrate how by using a network that allows features from the two streams to be jointly back propagated, we can achieve improved action recognition performance.

%
   
\section{Action Recognition Model}

 In our action recognition framework we extract features from each video, frame wise, via a Convolutional Neural Network (CNN) and pass those features, sequentially through an LSTM in order to classify the sequence. We introduce four CNN to LSTM fusion methods and evaluate their strength and weaknesses for recognising human actions. The architectures vary when considering how the features from the CNN model are fed to the LSTM action classification model. In this paper we are considering two direct mapping models as well as two merged models. The CNN and LSTM architectures we employ are discussed in Sections 3.1 and 3.2 respectively, the fusion methods are discussed in detail in Section 3.3.
   
\subsection{Convolutional Neural Network (CNN)}

\begin{figure}[h]
        \begin{flushleft}
        	\includegraphics[width=0.48\textwidth]{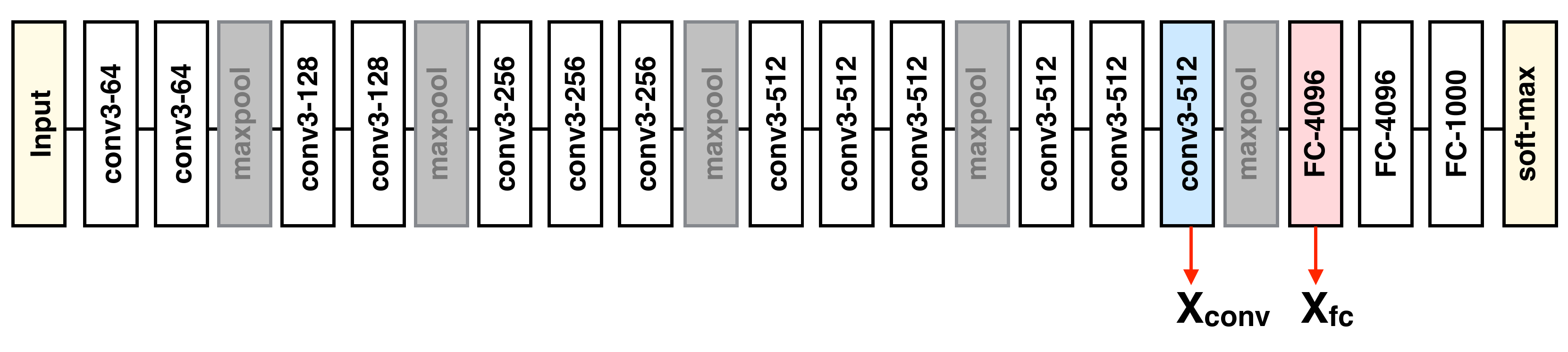}
	\caption{ VGG-16 Network \cite{VGG16}  Architecture : The network has 13 convolutional layers, \textbf{conv$<$receptive\_field\_size$>$-$<$number\_of\_channels$>$}, followed by 3 fully connected layers, \textbf{FC\-$<$number\_of\_channels$>$}. We extract features from the last convolutional layer and the first fully connected layer.}
	\label{fig:vgg16}
        \end{flushleft}
        \vspace{-2 mm}
  \end{figure}

   In our model a convolutional neural network has been used to perform the task of automatic feature learning. Most of the other traditional learning approaches accept a limited amount of training examples where beyond that limit no improvement can be observed. However, deep networks achieve higher performance with increasing amounts of training data. Therefore it requires a tremendous amount of data to train a CNN from scratch to gain higher performance. For the task of human action recognition it is more challenging to train a CNN from scratch with currently available datasets. For an instance, in UCF11 we only have around 150 videos per action class. Motivated by \cite{Simonyan2014} and in order to overcome the problem of limited training, we have used a pre-trained CNN model (ILSVRC-2014 \cite{VGG16}) which is trained with the ImageNet database. The network architecture is shown in Figure \ref{fig:vgg16}. It consists of 13 convolutional layers followed by 3 fully connected layers. In Figure \ref{fig:vgg16} we followed the same notations as \cite{VGG16} , where convolutional layers are represented as conv$<$receptive\_field\_size$>$ - $<$number\_of\_channels$>$ and  fully connected layers as FC-$<$number\_of\_channels$>$.  In the proposed action recognition model we use hidden layer activations from two hidden layers, namely the last convolutional layer ($X_{conv}$) and the first fully connected layer ($X_{fc}$). In this work we have initialised the network with weights obtained by pre-training with ImageNet, and fine tuned all the layers in order to adapt the network to our research problem. 
   
   \begin{figure}[h]
        \begin{flushleft}
        	\includegraphics[width=0.48\textwidth]{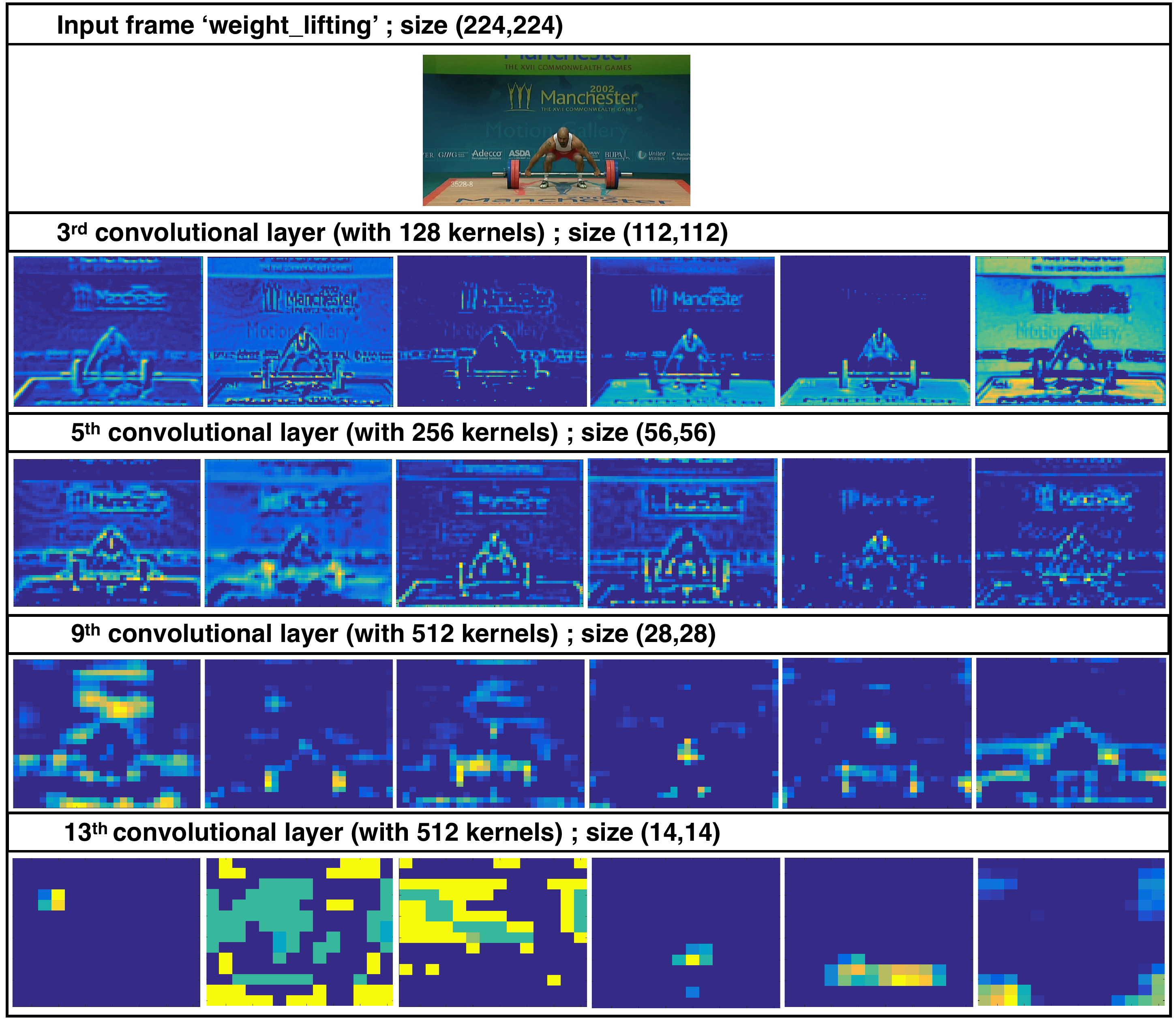}
	\caption{Heat maps of Intermediate CNN layer outputs for the video frame obtained from UCF Sports dataset :  outputs are obtained for the $3^{rd}$, $5^{th}$, $9^{th}$ and $13^{th}$ convolutional layers and only the first 6 channels for each layer is shown in this figure. }
	\label{fig:intermid}
        \end{flushleft}
        \vspace{-7 mm}
  \end{figure}

   Figure \ref{fig:intermid} shows an exemplar frame obtained from a video sequence of the UCF sports database and several corresponding intermediate CNN layer activations. The heat map values illustrates that with the fine tuning process the network has learned that when recognising the `weight lifting' action category most relevant information comes from athletes body (as they provide motion information), the stage floor and the equipment used. In the deeper layers these extracted salient features are combined and compressed in order to provide a rich input feature to the LSTM network.
   
   
\subsection{LSTM Network}

Recurrent neural networks are a type of artificial neural network well suited to handling sequential data. The main difference comes with the architecture of the network where the units are connected by directed cycles. Conventional recurrent neural networks such as Back Propagation Through Time (BPTT) \cite{BPTT1992} or Real Time Recurrent Learning (RTRL) \cite{RNN1989} are incapable of handling long time dependencies in practical situations due to the problem of vanishing or exploding gradients \cite{vanishGrad1994}. However, Hochreiter et al.\cite{LSTM1997} introduced an improved recurrent network architecture called Long-Short Term Memory (LSTM), which is composed of an appropriate gradient based learning algorithm. This architecture is capable of overcoming most of the fundamental problems of the conventional recurrent networks and is widely used in the area of computer vision and machine learning for modelling long term temporal correspondences within feature vectors.

\subsection{Combining CNNs and LSTMs for Action Recognition}

 In this study, we introduce four models that are capable of recognising human actions in challenging video sequences. The first two models (i.e. \textit{conv-L} and \textit{fc-L}) are single stream models based on extracting CNN activation outputs from the last convolution layer and the first fully connected layer respectively for each selected frame (see Section 4.2) of each video. Then the extracted CNN features are fed to an LSTM network and final output is decided by considering the  output of the soft-max layer. The second two approaches, \textit{fu-1} and \textit{fu-2}, are based on the idea of merging the two networks. Here we highlight the importance of the two stream approach where joint back propagation between streams is plausible. Exact model structures are described in the following four subsections.

\subsubsection{Convolutional-to-LSTM (\textit{conv-L})}
 
 In this model we take the input from the last convolutional layer of the VGG-16 network and feed it to the LSTM model. Let the last convolutional layer output for the $i^{th}$ video sequence be,
  
\begin{equation}
x^i_{conv}=[x^{i,1}_{conv}, x^{i,2}_{conv}, \ldots,x^{i,T}_{conv}],
\label{eq:1}
\end{equation}

where, T is the number of frames in the shortest video of the database.
Then the LSTM output can be defined as, 

\begin{equation}
h^i_{conv}= LSTM(x^i_{conv}).
\label{eq:2}
\end{equation}

The respective classification is given by, 

\begin{equation}
y^i=softmax(h^i_{conv}).
\label{eq:3}
\end{equation}

\subsubsection{Fully connetced-to-LSTM (\textit{fc-L})}

Similar to the \textit{conv-L} model in Section 3.3.1, equations for fully connected layer output for the $i^{th}$ video sequence with T frames can be written as follows,

\begin{equation}
x^i_{fc}=[x^{i,1}_{fc}, x^{i,2}_{fc}, \ldots,x^{i,T}_{fc}],
\label{eq:4}
\end{equation}

where $x^i_{fc}$ is the fully connected layer output for the $i^{th}$ video sequence, 

\begin{equation}
h^i_{fc}= LSTM(x^i_{fc}),
\label{eq:5}
\end{equation}

\begin{equation}
y^i=softmax(h^i_{fc}).
\label{eq:6}
\end{equation}

\subsubsection{\textit{Conv-L} + \textit{fc-L} fusion method 1 (\textit{fu-1} )}

In our third model we merge the outputs of Equations \ref{eq:2} and \ref{eq:5} and pass them through a soft-max layer to evaluate the final classification. This can be represented as,

\begin{equation}
y^i=softmax( W [h^i_{conv},h^i_{fc}] ).
\label{eq:7}
\end{equation}

\begin{figure}[t!]
    \centering
    \begin{subfigure}{.45\columnwidth}
        \includegraphics[width=.95\columnwidth]{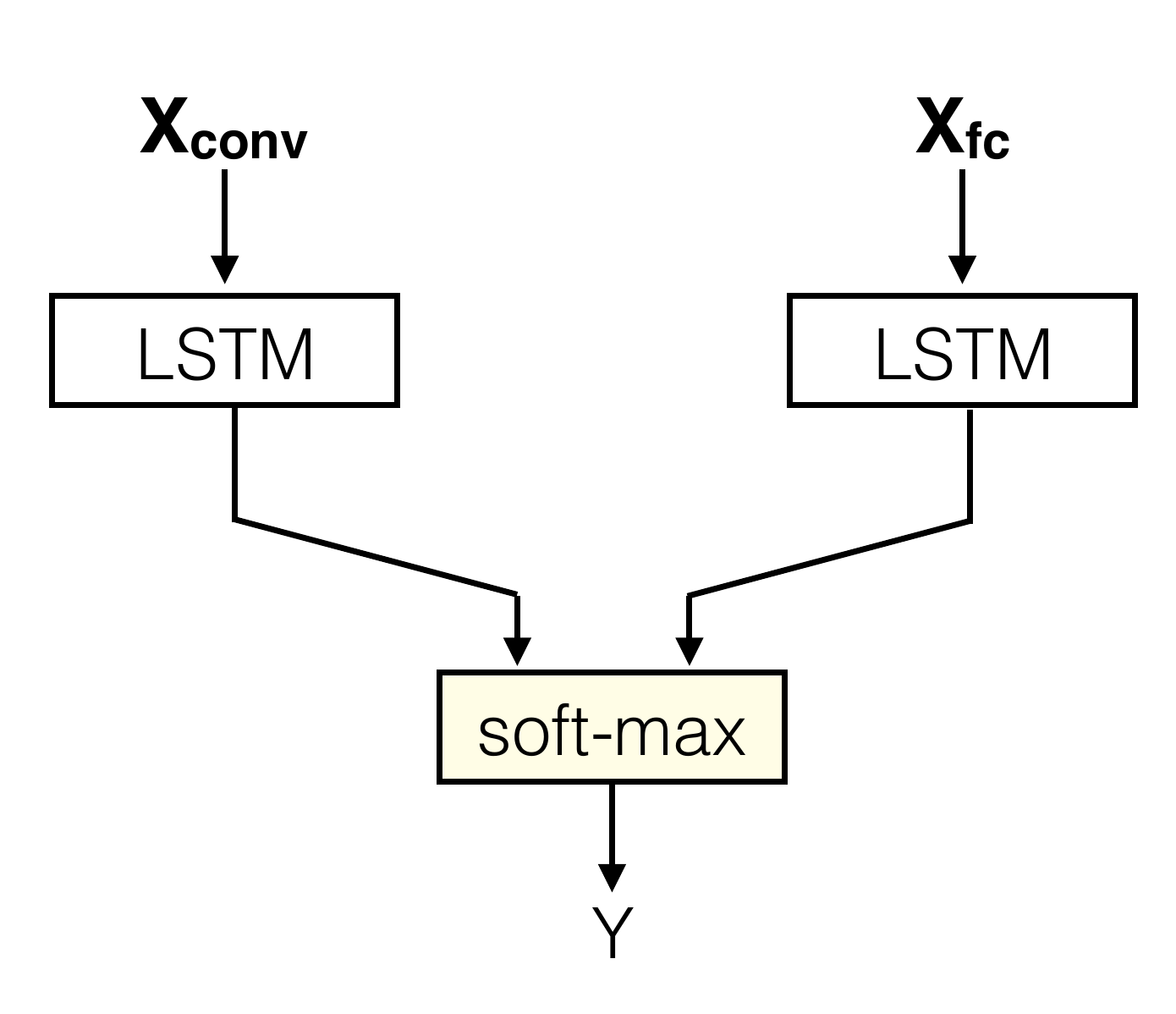} %
        \caption{\textit{fu-1}}
    \end{subfigure}
    \begin{subfigure}{.45\columnwidth}
        \includegraphics[width=.95\columnwidth]{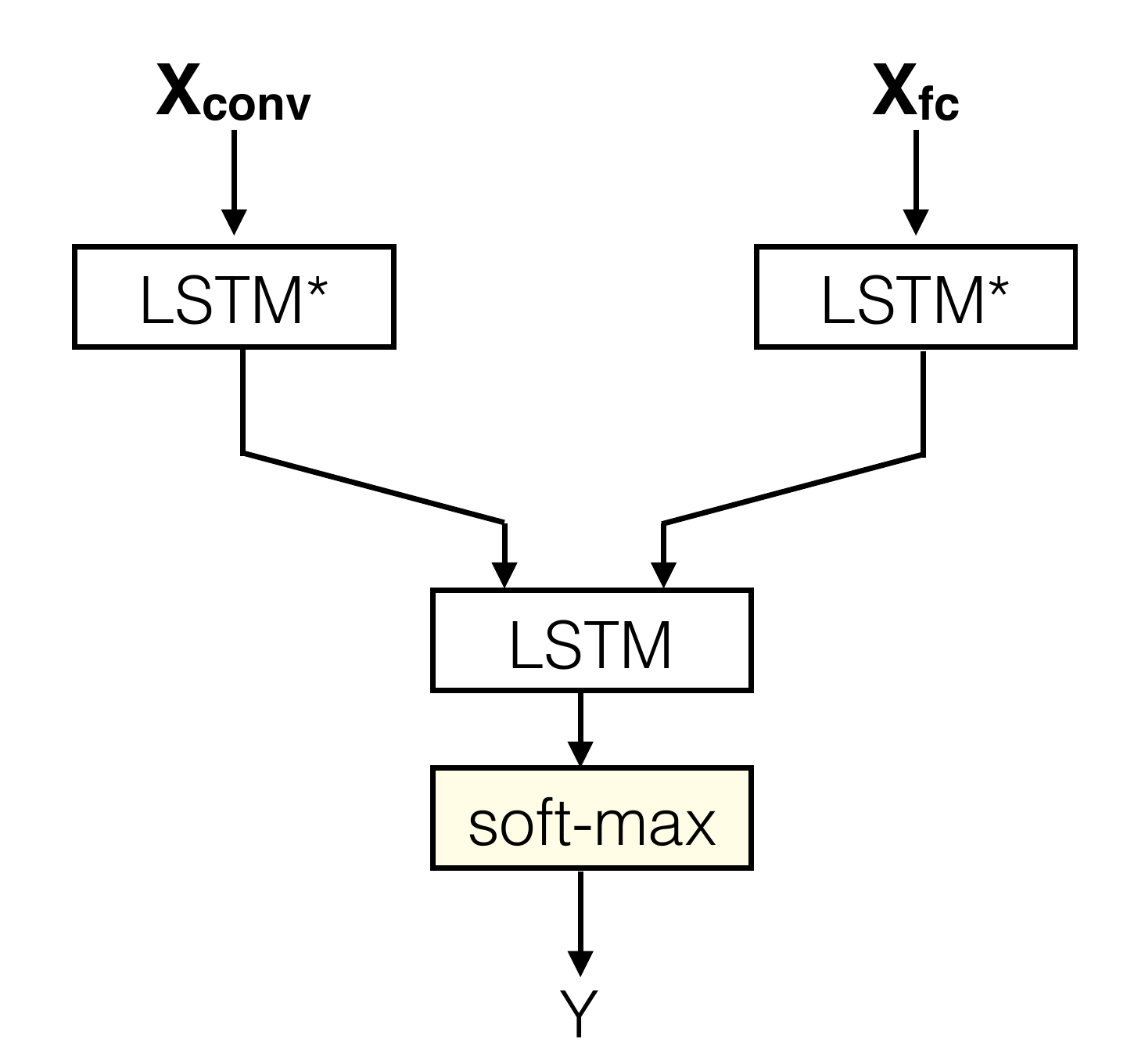}
        \caption{\textit{fu-2}}
         \end{subfigure}
         \vspace{-3 mm}
         \caption{Simplified diagrams of last two models \textit{fu-1} (left) and \textit{fu-2} (right)} 
          \label{fig:models}
    \begin{subfigure}{.9\columnwidth}
        \includegraphics[width=.95\columnwidth]{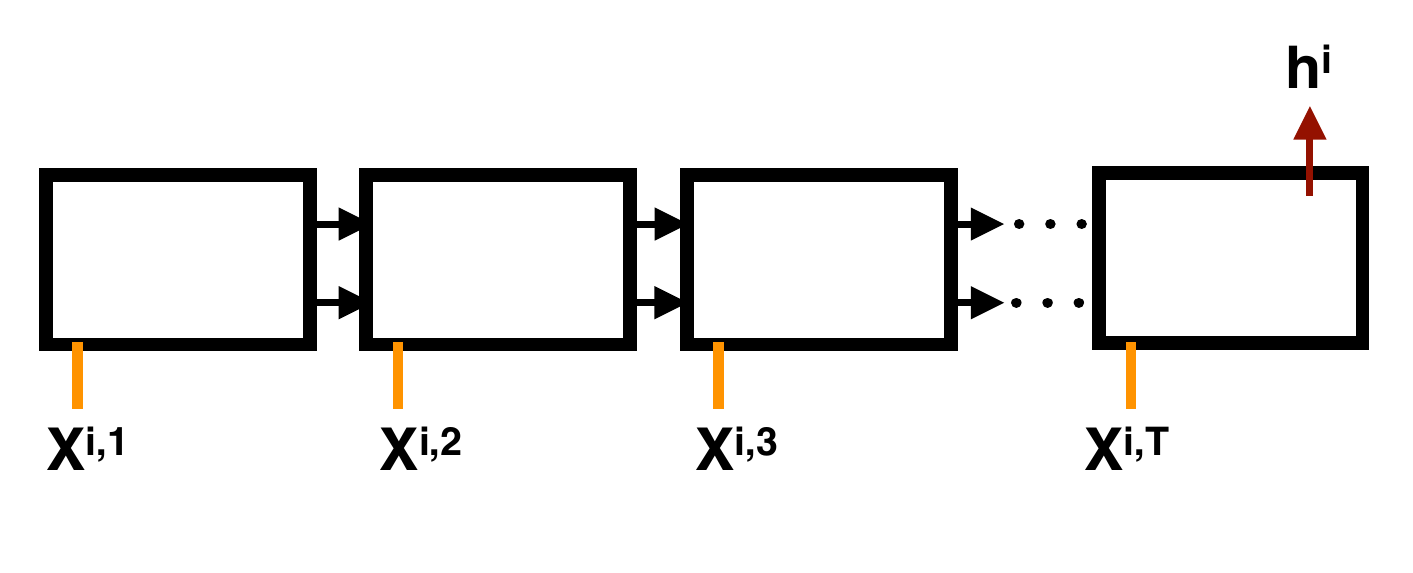} %
        \vspace{-5 mm}
        \caption{Sequence-to-One LSTM}
    \end{subfigure}
    \begin{subfigure}{.9\columnwidth}
        \includegraphics[width=.95\columnwidth]{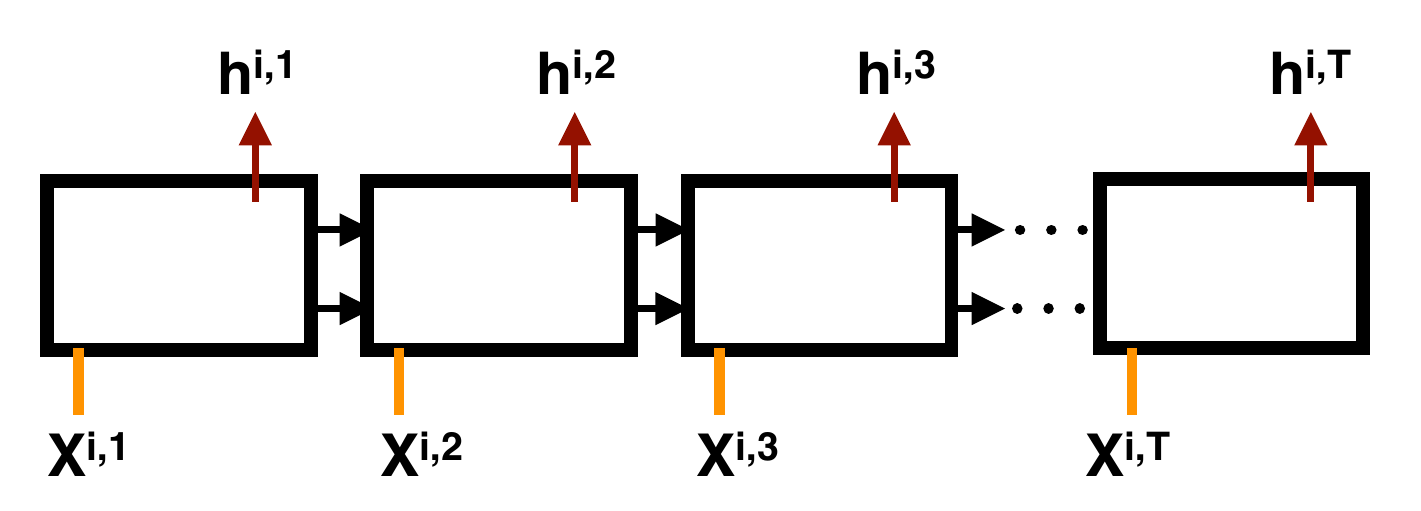}
        \vspace{-2 mm}
        \caption{Sequence-to-Sequence LSTM*}
         \end{subfigure}

         \caption{Two types of LSTMs : Sequence-to-One (LSTM) and Sequence-to-Sequence (LSTM*) are used with our last model. }
        \label{fig:lstms}
\end{figure}

The architecture is shown in Figure \ref{fig:models} (a).

\subsubsection{\textit{Conv-L} + \textit{fc-L} fusion method 2 (\textit{fu-2})}

In the above 3 models, each video sequence is represented as a single hidden unit (i.e. $h^i_{conv}$ , $h^i_{fc}$). In contrast, \textit{fu-2} represents each video sequence as a sequence of hidden states with the aid of a sequence-to-sequence (see Figure \ref{fig:lstms} (b)) LSTM. Finally those produced hidden states are passed through another LSTM which generates one single hidden unit representing that entire video (see Figure \ref{fig:models} (b)). The motivation behind having multiple layers of LSTMs is to capture information in a hierarchical manner. We carefully craft this architecture enabling joint back propagation between streams. Via this approach each stream (i.e. \textit{fc-L} and \textit{conv-L} ) is capable of communicating with each other and improving the back propagation process.

 Therefore, in this model equations for $h^i_{conv}$  and $h^i_{fc}$ are modified as shown below. The LSTMs that handle data in a sequence-to-sequence manner are represented as $LSTM^*$.

\begin{equation}
h^{i,t}_{conv}=LSTM^*(x^{i,t}_{conv}, h^{i,t-1}_{conv} ),
\label{eq:8}
\end{equation}

\begin{equation}
h^{i}_{conv}=[h^{i,1}_{conv},h^{i,2}_{conv}, \ldots,h^{i,T}_{conv}],
\label{eq:9}
\end{equation}

\begin{equation}
h^{i,t}_{fc}=LSTM^*(x^{i,t}_{fc}, h^{i,t-1}_{fc} ),
\label{eq:10}
\end{equation}

\begin{equation}
h^{i}_{fc}=[h^{i,1}_{fc},h^{i,2}_{fc}, \ldots,h^{i,T}_{fc}].
\label{eq:11}
\end{equation}

Then the resultant sequence of predictions are fed to the final LSTM which works in a sequence-to-one (see Figure \ref{fig:lstms} (a)) manner,

\begin{equation}
h^{i}= LSTM(W [h^i_{conv},h^i_{fc}]),
\label{eq:12}
\end{equation}

\begin{equation}
y^i=softmax(h^{i}).
\label{eq:13}
\end{equation}

\section{Experiments}

Our approach has been evaluated with three publicly available action datasets: UCF11 (YouTube action dataset) \cite{UCF11wild}, UCF Sports \cite{UCFSports2008, UCFSports2014}  and jHMDB \cite{jHMDB2013}; and the performance has been compared with state-of-the-art methods for each database.

\subsection{Datasets}

UCF11 can be considered a challenging dataset due to large variation of illumination conditions, camera motion and cluttered backgrounds that are present with the video sequences. It contains 1600 videos and all videos are acquired at a frame rate of 29.97 fps. The dataset represents 11 action categories : basketball shooting, biking/cycling, diving, golf swinging, horse back riding, soccer juggling, swinging, tennis swinging, trampoline jumping, volleyball spiking, and walking with a dog.

UCF sports consists of 150 video sequences categorised into 10 action classes; diving, golf swing, kicking, lifting, riding horse, running, skate boarding, swing bench, swing side and walking. The videos have a minimum length of 22 frames and maximum length of 144 frames. 

jHMDB contains 923 videos (with length ranging from 18 to 31 frames) belonging to 21 action classes. This dataset is an interesting and challenging data set which includes activities from daily life such as `brushing hair', `pour', `sit' and `stand' other than sports activities. 

\subsection{Experimental Setup}

Different databases provide videos of varying lengths. Therefore, in each database we set T to be the number of frames that are available in the shortest video sequence in that particular database. Extracting out the first T frames from a particular video is not optimal. Therefore we sample T number of frames in the following manner such that they are evenly spaced through the video.  

Let N be the total number of frames of the given video sequence. Then the variable $t$ can be defined as,

\begin{equation}\label{eq:1}
t=\cfrac{N}{T}.
\end{equation}
 
 Let $t'$ be the value after rounding $t$ towards zero (i.e. the integer part of $t$, without the fraction digits). Finally the subsequence $S$ with T frames is selected as,

\begin{equation}\label{eq:1}
S=[1\times{t'}, 2\times{t'},3\times{t'},\ldots,T\times{t'}].
\end{equation}

The training set for each database is used for fine tuning the VGG-16 network and training the LSTM network. Using VGG-16, we have input shapes of $X_{conv}=(512\times{14\times{14}})$ and $X_{fc}=(4096\times{1})$ which are the inputs to the LSTM network. In our setup, all the LSTMs have a hidden state dimensionality of 100. In order to minimise the problem of overfitting we have used a drop out ratio of 0.25 between LSTM output and the soft-max layer in all four models and also between the LSTMs in model \textit{fu-2}. The optimal number of hidden states and drop out ratio are determined experimentally. 

In experiments for the dataset UCF11, we have used Leave one out cross validation (LOOCV) as in the original work \cite{UCF11wild}. Experimental results for the UCF sports dataset have also been obtained by following the LOOCV scheme. The jHMDB dataset has been evaluated by considering the training and testing splits as in \cite{jHMDB2013} and the overall performance is measured by considering the average performance of the three splits.

\begin{table*}
\begin{center}
\begin{tabular}{|c|c||c|c||c|c|}
 \hline
	
      \multicolumn{2}{|c||}{UCF 11} &
      \multicolumn{2}{c||}{UCF Sports} &
      \multicolumn{2}{c|}{jHMDB} \\
  \hline  
  
       Method & Accuracy & Method & Accuracy & Method & Accuracy  \\
  \hline
	Hasan et al. \cite{Hasan2014} & 54.5\% & Wang et al. \cite{Wang2009} & 85.6\% &  Lu et al. \cite{Lu2015} & 58.6\%  \\	
	Liu et al. \cite{UCF11wild}& 71.2\% & Le et al. \cite{Le2011} & 86.5\% &  Gkioxari et al.\cite{actiontubes}  & 62.5\%\\	
	Ikizler-Cinbis et al \cite{Ikizler-Cinbis2010} & 75.2\% & Kovashka et al. \cite{Kovashka2010} & 87.2\% &  Peng et al. \cite{peng2014}  & 62.8\% \\
	Dense Trajectories \cite{densetrajectories} & 84.2\% & Dense Trajectories  \cite{densetrajectories} & 89.1\% & Jhuang et al. \cite{jHMDB2013} & \textbf{69.0}\% \\
	Soft attention \cite{Sharma2015} &84.9\%  & Weinzaepfel et al. \cite{weinzaepfel2015}& 90.5\% &  Peng et al. \cite{PengX2014}  & \textbf{69.0}\% \\
	Cho et al.\cite{Cho2014} & 88.0\%  & SGSH \cite{Abdulmunem2016}& 90.9\% &  &   \\	
	 Snippets \cite{Ravanbakhsh15} & 89.5\% & Snippets \cite{Ravanbakhsh15} & 97.8\% &   &   \\
\hline
\hline
       \textit{conv-L}  & 89.2\% & \textit{conv-L} & 92.2\%  & \textit{conv-L} & 52.7\%\\	
       \textit{fc-L}  & 93.7\% & \textit{fc-L} & 98.7\%  & \textit{fc-L} & 66.8\%\\	
       \textit{fu-1}  & 94.2\% & \textit{fu-1} & 98.9\%  & \textit{fu-1} & 67.7\%\\	
       \textit{fu-2}  & \textbf{94.6}\% & \textit{fu-2} & \textbf{99.1}\%  & \textit{fu-2} & \textbf{69.0}\%\\

\hline			
			
\end{tabular} 
\end{center}
\vspace{-3 mm}
\caption{Comparison of our results to the state-of-the-arts on action recognition datasets UCF Sport , UCF11 and jHMDB }\label{tab:tab_1}
\end{table*}

\subsection{Experimental Results}

We evaluated all 4 proposed methods against the state-of-the-art baselines and the results are tabulated in Table \ref{tab:tab_1}. In the UCF11 dataset and the UCF sports dataset, the proposed \textit{fc-L}, \textit{fu-1} and \textit{fu-2} has out performed the current state-of-the-art models and in jHMDB dataset our \textit{fu-2} has achieved the same results as the current state-of-the-art.

When comparing results produced by proposed 4 models, \textit{fu-2} produces the best accuracy value showing that the proposed fusion method yields best results with the aid of its deep layer wise structure. This can be considered the main reason for the improvement of the accuracy from \textit{fu-1} to \textit{fu-2} in all three databases.

\begin{figure}[t!]
    \centering
     \begin{subfigure}{.9\columnwidth}
      	  \includegraphics[width=.95\columnwidth]{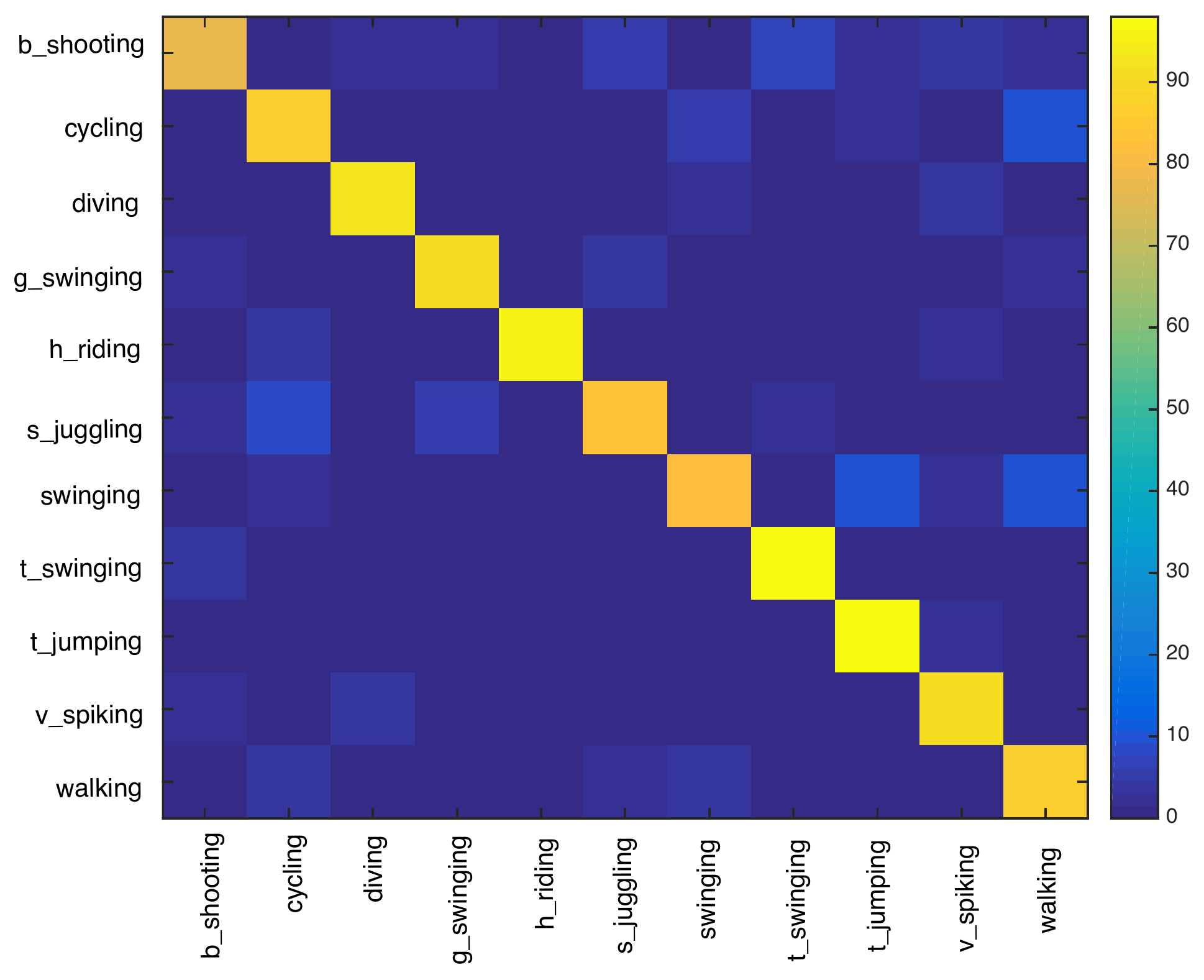} %
       	 \caption{Confusion matrix of \textit{conv-L}}
    \end{subfigure}
    \begin{subfigure}{.9\columnwidth}
        \includegraphics[width=.95\columnwidth]{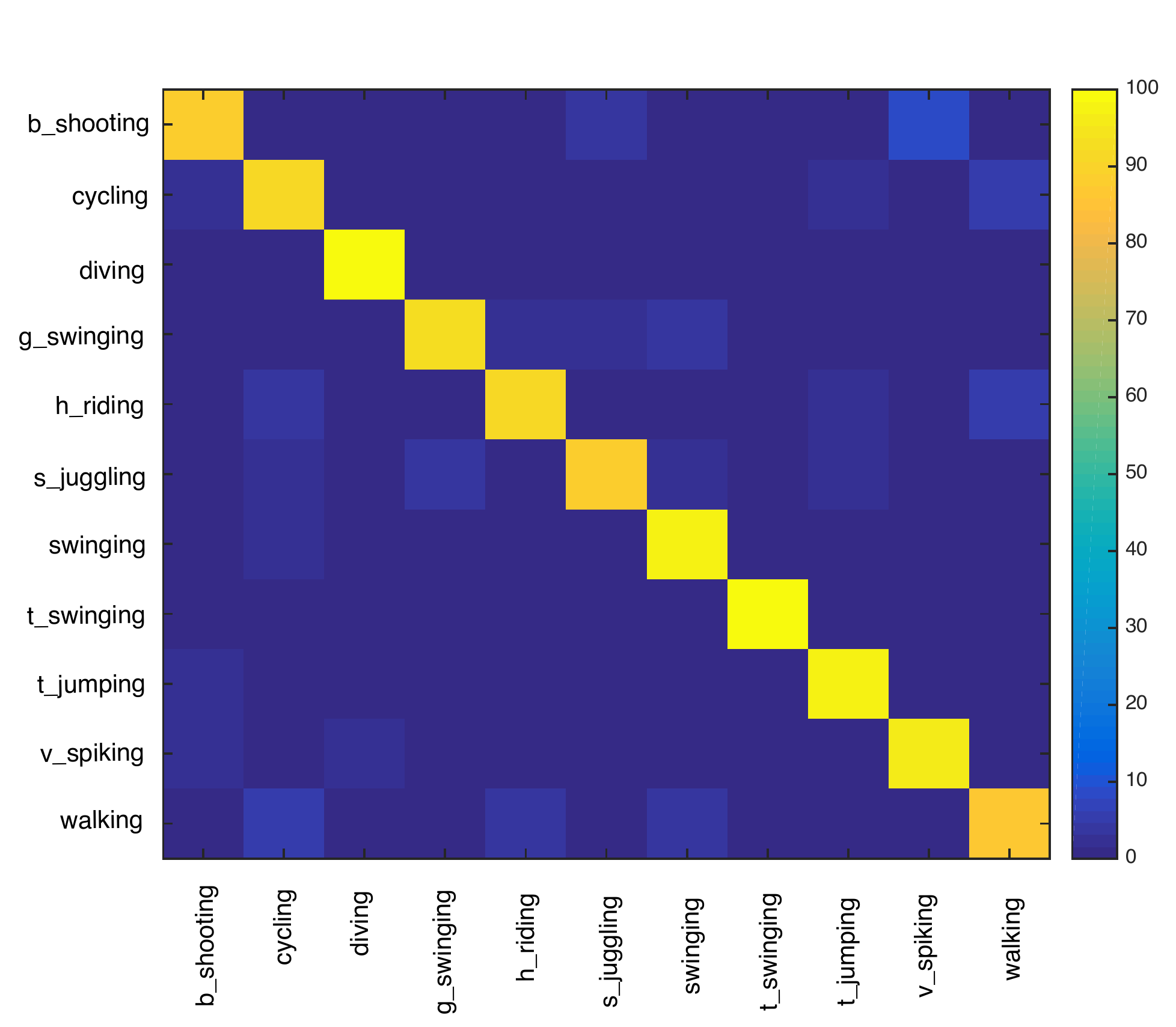}
        \caption{Confusion matrix of \textit{fc-L}}
     \end{subfigure}
     \vspace{-2 mm}
     \caption{Confusion matrices for the models \textit{conv-L} (a) and \textit{fc-L} (b) on UCF11 dataset with 11 action classes } 
     \label{fig:conf_mat_fortwo}
     \vspace{-5 mm}
    \end{figure}

Direct mapping methods (i.e. \textit{conv-L} and \textit{fc-L}) have less accuracy compared to merged models yet produce highly competitive results when compared to the baselines. The model \textit{conv-L} has the lowest accuracy due to the highly dense features from the convolutional layer are difficult to discriminate. This can be considered the main reason for the difference between accuracies in \textit{conv-L} and \textit{fc-L}. We have observed that \textit{fc-L} is prone to confusing target actions with more classes compared to the model \textit{conv-L} for most of the action classes. For example, in the results for UCF11 for the action class `basketball\_shooting' model \textit{conv-L} misclassifies the action as `diving', `g\_swing', `s\_juggling' , `t\_swinging', `t\_jumping', `volleyball\_spiking' and `walking' (see Figure \ref{fig:conf_mat_fortwo} (a));  while the model  \textit{fc-L} has confusions only with `s\_juggling', `t\_swinging' and `volleyball\_spiking' (see Figure \ref{fig:conf_mat_fortwo} (b)). 

We observed that the convolution layer output contains more spatial information and the fully connected layer output contains more discriminative features among those spatial features, that provides vital clues when discriminating between those action categories. Therefore, from the experimental results it is evident that more spatiallly related information such as objects (e.g. ball) or sub action units (i.e. basketball\_shooting can be considered a combination of sub actions including running, jumping, and throwing) contained within the main action can confuse the model. We also note that the action `horse\_riding' has been well classified by the model \textit{conv-L}  when compared with the performance of the \textit{fc-L} model. An action such as `horse\_riding' can be accurately classified with the spatial information as it is always composed of the `horse' and the `rider'. Therefore, this is further evidence to demonstrate the convolutional layer output is composed of more spatial information. 

%

We overcome the above stated drawback in the fusion models \textit{fu-1} and \textit{fu-2}, where the  models identify that the fully connected layer output represents the main information queue while the convolutional output represents the complementary information such as spatial information from the background. 
When drawing parallels to the current state-of-the-art techniques, in both \cite{Baccouche2011} and \cite{Sharma2015} authors extract out the final convolutional layer output from the CNN and pass it through a single LSTM stream to generate the related classification. The main distinction between \cite{Baccouche2011} and \cite{Sharma2015} is that the latter has an additional attention mechanism which tells which area in the feature vector to look at. We believe that when packing such highly dense input features together and providing it directly to a LSTM model, the model may get ``confused'' regarding what features to look at. The attention mechanism in \cite{Sharma2015} can be considered a mechanism to improve the situation yet it has not been capable of fully resolving the issue (see row 5, column 1 in Table \ref{tab:tab_1} ). \par

From the experimental results it is evident that the fully connected layer features posses more discriminative power than the convolutional layer features, yet are a sparse representation. Therefore in our fourth model \textit{fu-2}, by providing the convolutional input and fully connected layer input to separate LSTM streams, the next LSTM layer (merged LSTM layer in Figure \ref{fig:models} b) is able to achieve a good initialisation from the \textit{fc-L} stream and then jointly back propagate and learn which areas of the \textit{conv-L} stream to focus on. Hence this deep layer wise fusion mechanism gains the state-of-the-art results with exemplary precision. We compare this against \textit{fu-1} where such joint back propagation is not possible. The back propagation in \textit{fc-L} and \textit{conv-L} streams happens separately and hence the model is mainly driven by the \textit{fc-L} stream. In contrast, the 2 streams in \textit{fu-2} model are coupled allowing them to provide complementary information to each other. This hypothesis is further verified by the lack of improvement in accuracy when moving from \textit{fc-L} to \textit{fu-1} when compared against the improvement between models \textit{fu-1} and \textit{fu-2}. 
\par
 In the following subsections we evaluate the results produced by our fourth model \textit{fu-2} with the current state-of-the-art, as it has produced the highest accuracy.

\subsubsection{Results on UCF11 dataset}

\begin{figure}[h]
        \begin{flushleft}
        	\includegraphics[width=0.45\textwidth]{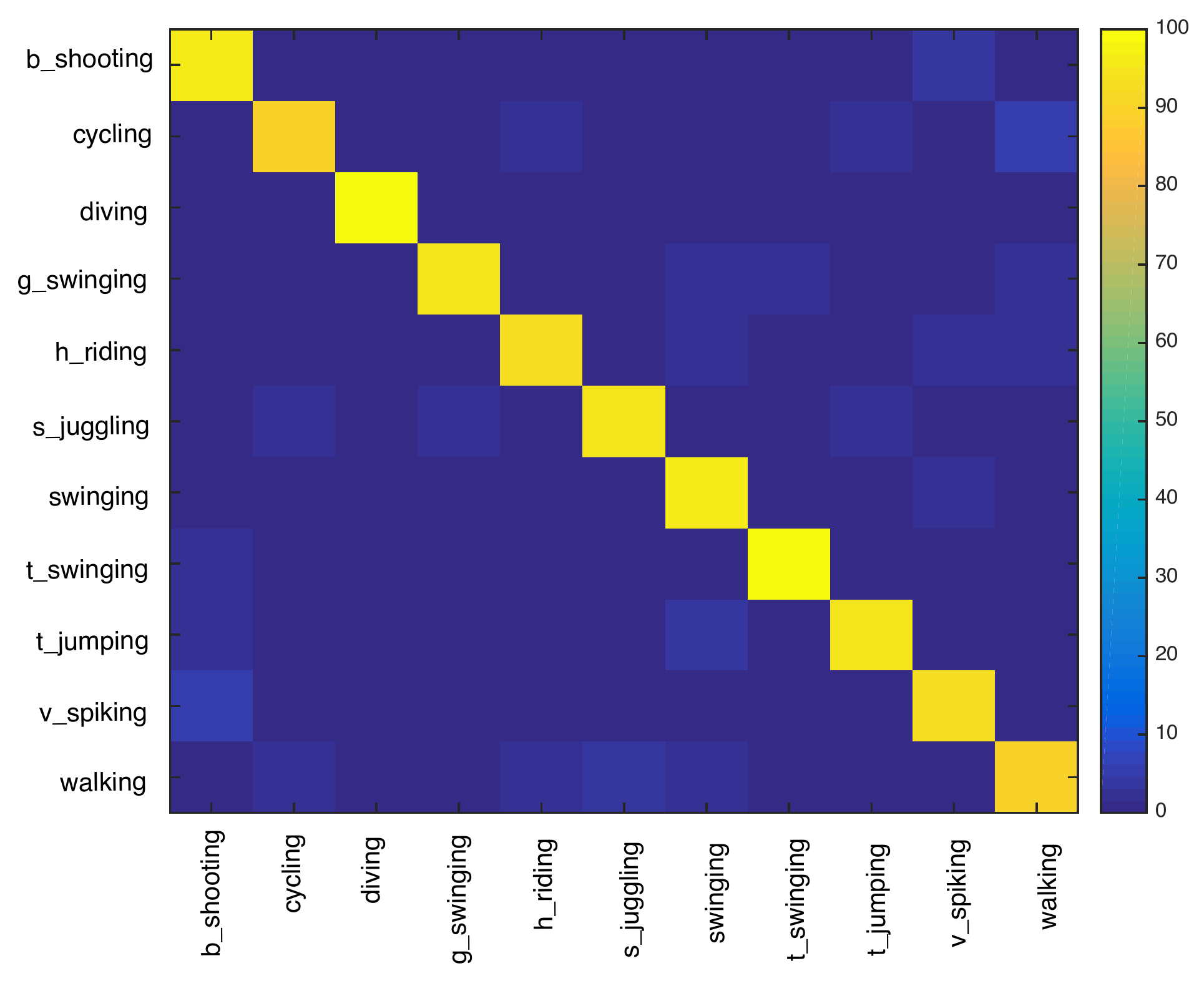}
	\vspace{-2 mm}
	\caption{Confusion matrix of proposed model \textit{fu-2} for UCF 11 dataset}
	\label{fig:fig5}
        \end{flushleft}
        \vspace{-3 mm}
  \end{figure}

 From the results, it is clear that our proposed method achieves excellent performance for each action class separately. The class `diving' has the highest accuracy of 99.98 while the `cycling' class shows the lowest accuracy of 89.69 which is still can be considered as reasonable performance. There are several minor confusions between classes such as volleyball spiking, basketball shooting and cycling, walking due to the fact that the confusing action classes present similar backgrounds and similar motions. Yet when drawing comparisons to other state of the art results (see column 1, Table \ref{tab:tab_1}) our proposed architecture is more accurate as it achieves an accuracy of 94.6, 5.1 percent higher than the previous highest result of \cite{Ravanbakhsh15}. Figure \ref{fig:fig5} depicts the confusion matrix of proposed model \textit{fu-2} for each action category in UCF11 dataset.
  
 \vspace{-7 mm}
 
 \subsubsection{Results on UCF Sports dataset}

  A recent study by Abdulmunem et al.\cite{Abdulmunem2016}, introduced the method SGSH to select only the points of interest (salient) regions of the video sequence discarding other information from the background. When handcrafting such background segmentation processes in video sequences with background clutter, camera motion and illumination variances, it tends to produce erroneous results. 
  
  In our approach the automatic feature learning in both CNN and LSTM networks has learnt the salient regions in the video sequences. In \cite{Abdulmunem2016} with the presented confusion matrix (see Figure 8 (a) in \cite{Abdulmunem2016}) it is observed that there is a confusion between action categories such as kicking with skating and golf swing with kicking and skate boarding; in contrast our model has achieved substantially higher accuracies when discriminating those action classes. 
  
\begin{figure}[!h]
   \vspace{-3 mm}
        \begin{flushleft}
        	\includegraphics[width=0.45\textwidth]{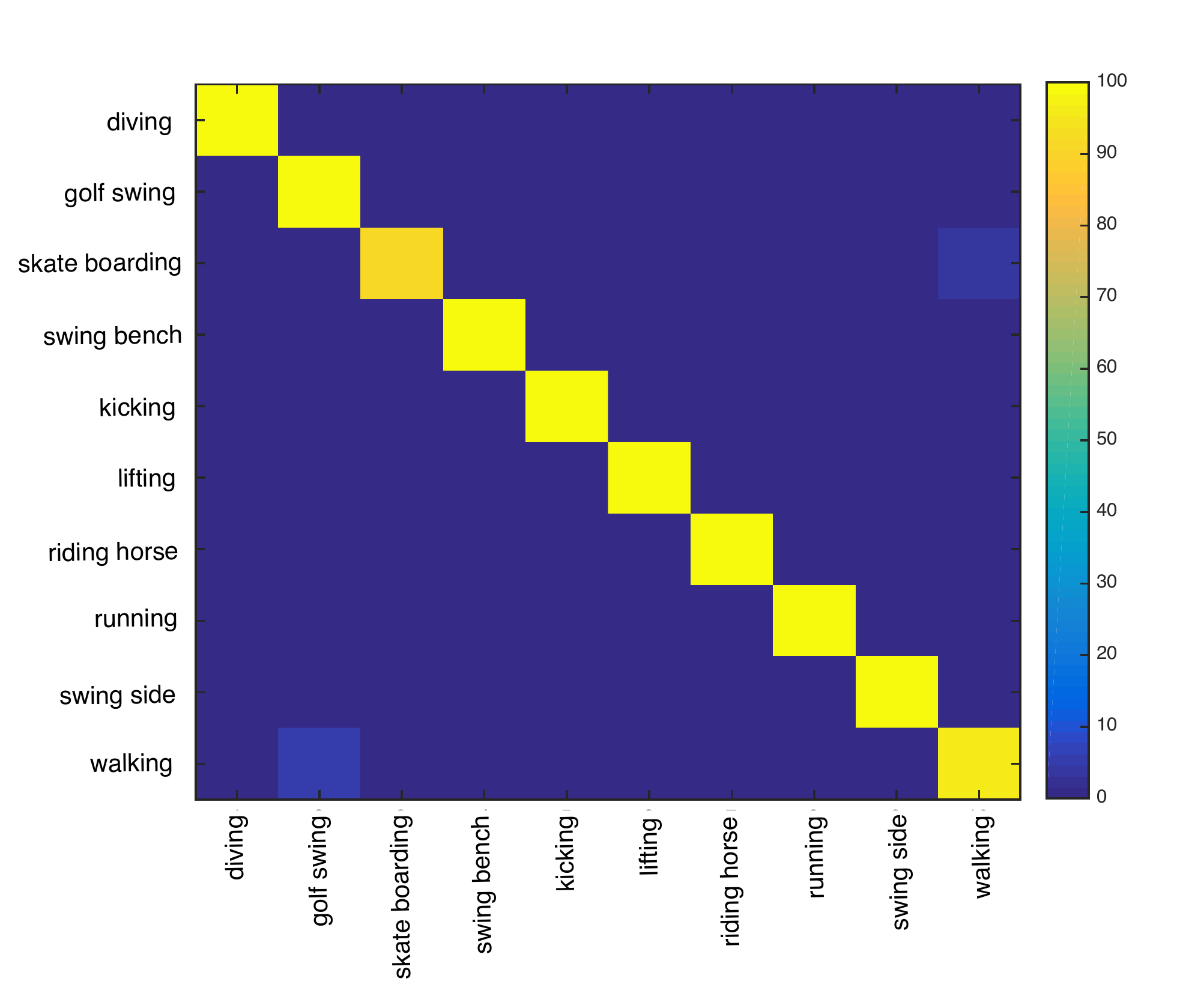}
	\vspace{-4 mm}
	\caption{Confusion matrix of our model \textit{fu-2} for UCF Sports dataset}
	\label{fig:fig6}
        \end{flushleft}
        \vspace{-1 mm}
  \end{figure}

  Figure \ref{fig:fig6} shows the accuracy results for each action category. According to the results on UCF sports data our approach has the ability to discriminate sports actions well.
  
\vspace{-7 mm}
 \subsubsection{Results on jHMDB dataset}

\begin{figure}[!h]
 \vspace{-2 mm}
        \begin{flushleft}
        	\includegraphics[width=0.49\textwidth]{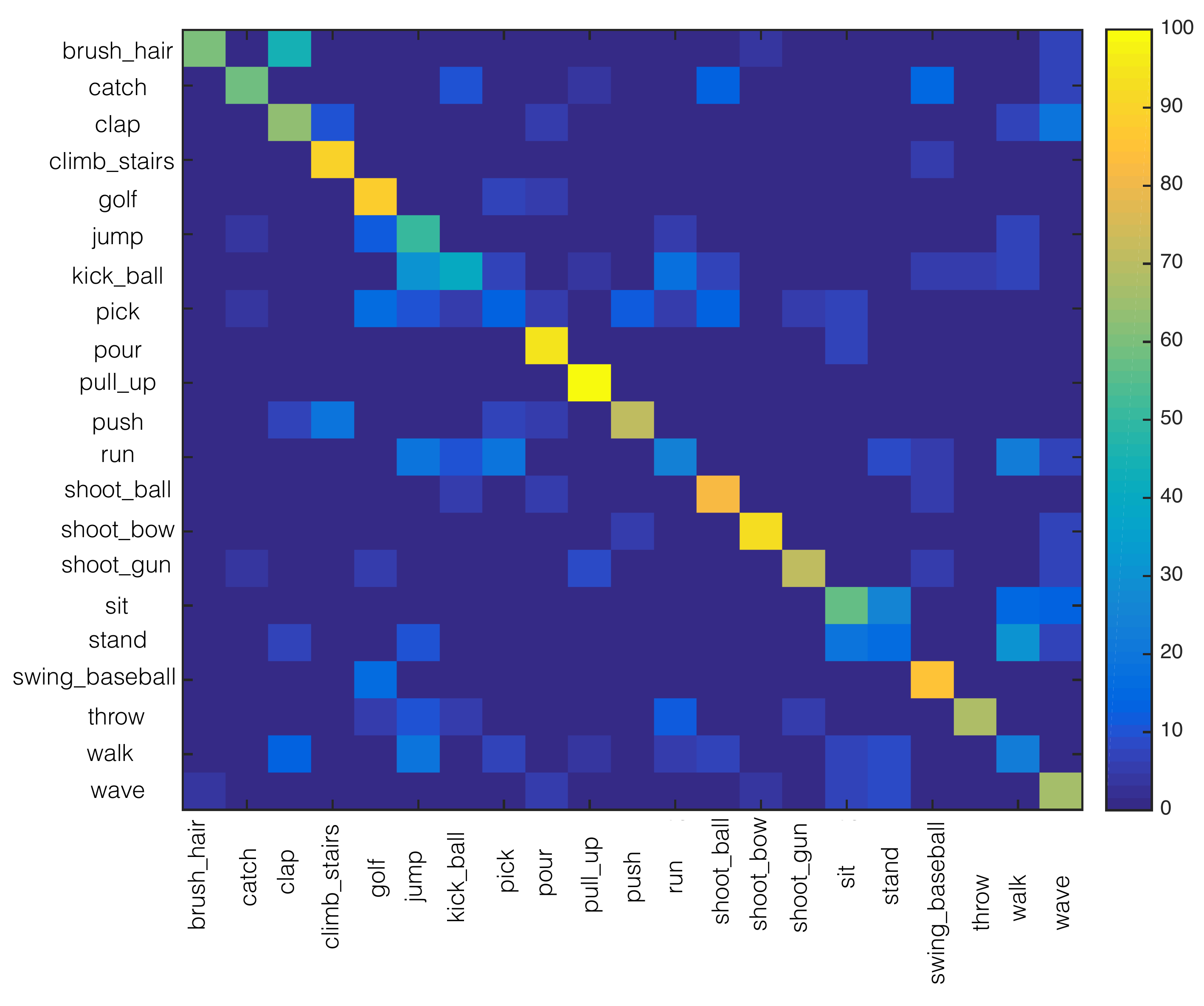}
	\vspace{-5 mm}
	\caption{Confusion matrix of our \textit{fu-2} for jHMDB dataset with 21 action classes}
	\label{fig:fig4}
        \end{flushleft}
        \vspace{-4 mm}
  \end{figure}

The confusion matrix of the proposed \textit{fu-2} architecture is shown in Figure \ref{fig:fig4}, and gives a clear idea regarding the discriminative ability of our method on different action categories. According to the results our \textit{fu-2} model has a high discriminative ability when considering the actions such as `pour', `golf', `climb\_stairs', `pull\_up' and `shoot\_ball'. However the actions such as `pick', `run', `stand' and `walk' shows lower accuracy values.  In the previous work \cite{actiontubes}, the accuracy for classes `shoot\_ball' and `jump' are considerably lower than the other action classes. However, in comparison to their work, our method shows higher ability to discriminate those action classes well. In \cite{actiontubes}, the authors have elaborated that motion features are more significant than spatial features when classifying action classes such as `clap', `climb\_stairs', `sit', `stand', and `swing\_baseball'. The motion features in their work were obtained via a hand-crafted optical flow feature. In contrast to their approach we do not explicitly model the motion features, but still the LSTM network maps the correlation between the spatial features in the input sequence in an automatically. This captures and models the motion and produces a significantly higher accuracy compared to \cite{actiontubes}.

 \vspace{-5 mm}  
In our proposed models, fu-1 and fu-2, the number of trainable parameters in the LSTM networks are 5.8M and 5.9M respectively. A further 134.3M parameters are present in the VGG-16 network which we fine tune. Thus, the total parameters in the models do not exceed 141M. Compared to other approaches, \cite{actiontubes} and \cite{Simonyan2014} both contain approximately 180M parameters. \cite{Jeff2015} and \cite{Ravanbakhsh15} are both derived from AlexNet and have approximately 300M parameters; all of which significantly exceeds the number of parameters in the proposed method. 

\section{Conclusion} 

  We proposed an approach which uses convolutional layer outputs for the training of an LSTM network for recognising human actions. We evaluate four fusion methods for combining convolutional neural network outputs with LSTM networks, and show that these methods outperform state of the art approaches for three challenging datasets. According to our study, the combination of two streams of LSTMs that use the final convolutional layer output and the first fully connected layer output respectively shows higher recognition ability than using either stream on its own. Further performance improvement is gained by adding a third LSTM layer for the purpose of combining the two streams. From the experimental results we illustrate that the fully connected layer output acts as an attention mechanism to direct the LSTM through the important areas of the convolutional feature sequence.

\section*{Acknowledgement}
This research was supported by an Australian Research Council (ARC) Linkage grant LP140100221.


{\small
\bibliographystyle{ieee}
\bibliography{paper1}
}

\end{document}